\documentclass[lettersize,journal]{IEEEtran}
\usepackage{amsmath,amsfonts}
\usepackage{algorithmic}
\usepackage{algorithm}
\usepackage{array}
\usepackage{textcomp}
\usepackage{stfloats}
\usepackage{url}
\usepackage{verbatim}
\usepackage{graphicx}
\usepackage{cite}
\usepackage{amssymb}
\usepackage{multirow}
\usepackage{booktabs}
\usepackage{graphicx}
\usepackage{subfigure}
\usepackage{color,xcolor}
\usepackage{hyperref}
\usepackage{threeparttable}
\usepackage{multirow}
\usepackage{amsmath}
\usepackage{makecell}

\begin{document}

\title{CF-YOLO: Cross Fusion YOLO for Object Detection in Adverse Weather with a High-quality Real Snow Dataset}
\author{Qiqi Ding, Peng Li,
        Xuefeng Yan, Ding Shi, Luming Liang, Weiming Wang, \textit{Member}, \textit{IEEE}, Haoran Xie, \textit{Senior Member}, \textit{IEEE}, Jonathan Li, \textit{Senior Member}, \textit{IEEE}, and  Mingqiang Wei, \textit{Senior Member}, \textit{IEEE}
\thanks{Q. Ding, P. Li, X. Yan, D. Shi and M. Wei are with the School of Computer Science and Technology, Nanjing University of Aeronautics and Astronautics, Nanjing, China (e-mail: qqding@nuaa.edu.cn, pengl@nuaa.edu.cn, yxf@nuaa.edu.cn, shiding@nuaa.edu.cn, mingqiang.wei@gmail.com).}
\thanks{L. Liang is with Microsoft, USA (llmpass@gmail.com).}
\thanks{W. Wang is with the School of Science and Technology, Hong Kong Metropolitan University, Hong Kong, China (wmwang@hkmu.edu.hk).}
\thanks{H. Xie is with the Lingnan University, Hong Kong, China (e-mail: hrxie@ln.edu.hk).}
\thanks{J. Li is with the Department of Geography and Environmental Management and Department of Systems Design Engineering, University of Waterloo, Waterloo, Canada (e-mail: junli@uwaterloo.ca).}


}

\markboth{Journal of \LaTeX\ Class Files,~Vol.~14, No.~8, August~2021}%
{Shell \MakeLowercase{\textit{et al.}}: A Sample Article Using IEEEtran.cls for IEEE Journals}


\maketitle

\begin{abstract}
Snow is one of the toughest adverse weather conditions for object detection (OD). Currently, not only there is a lack of snowy OD datasets to train cutting-edge detectors, but also these detectors have difficulties learning latent information beneficial for detection in snow. To alleviate the two above problems, we first establish a real-world snowy OD dataset, named RSOD. Besides, we develop an unsupervised training strategy with a distinctive activation function, called $Peak \ Act$, to quantitatively evaluate the effect of snow on each object. Peak Act helps grading the images in RSOD into four-difficulty levels. To our knowledge, RSOD is the first quantitatively evaluated and graded snowy OD dataset.
Then, we propose a novel Cross Fusion (CF) block to construct a lightweight OD network based on YOLOv5s (call CF-YOLO).
CF is a plug-and-play feature aggregation module, which integrates the advantages of Feature Pyramid Network and Path Aggregation Network in a simpler yet more flexible form.
Both RSOD and CF lead our CF-YOLO to possess an optimization ability for OD in real-world snow. 
That is, CF-YOLO can handle unfavorable detection problems of vagueness, distortion and covering of snow.
Experiments show that our CF-YOLO achieves better detection results on RSOD, compared to SOTAs.
The code and dataset are available at \textcolor{magenta}{ \href{https://github.com/qqding77/CF-YOLO-and-RSOD}{https://github.com/qqding77/CF-YOLO-and-RSOD}}.
\end{abstract}


\section{Introduction}

\IEEEPARstart{C}{NN}-based detectors heavily depend on the integrity of objects in an image \cite{Hnewa21,HuangLJ21,tgrs/GongXTSXDL20}. Unfortunately, objects are often partially or even fully covered by snow in winter. Images captured under such adverse weather will spontaneously lose the significant information of describing objects, thus easily leading to the collapse of most object detection (OD) methods which originally behave well in normal weather \cite{DBLP:journals/corr/RenHG015,tgrs/ZhangYFL19,DBLP:journals/tgrs/GuoYYTHL21a}.

There exist two major challenges in detecting objects covered by snow: (1) Capturing snow/snow-free image pairs in real-world scenarios is extremely difficult or nearly impossible.
Therefore, existing detectors will train themselves on benchmark datasets either captured under normal weather conditions \cite{DengDSLL009,DBLP:conf/eccv/LinMBHPRDZ14,DBLP:journals/tgrs/CaoFXM22,DBLP:journals/tgrs/ZhangLZ19}  or by synthetic snow (e.g., SnowCityScapes \cite{DBLP:journals/corr/abs-2103-11298}). Such trained models have the well-known domain shift problem, thus generalizing poorly in real-world snowy scenarios.
(2) Existing detectors have difficulties learning latent information beneficial for detection in snow, since snow can destroy the low-level vision information (texture, outline, etc.) of the objects in the images.

In this paper, we observe that the performance of current vision techniques is still mainly benchmarked under normal weather conditions. Even the top-performing object detectors undergo severe performance degradation under adverse weather conditions. Therefore, we raise a practically meaningful detection question under adverse weather: 
Does the synergy of establishing a real-snow OD dataset and developing a feature aggregation module to learn latent information, actually enhance the capability of cutting-edge OD networks in the snowy condition?

To answer this question, (1) we collect a high-quality outdoor dataset (RSOD) towards real-snow object detection. RSOD contains 2100 real-world snowy images annotated in the format of COCO and YOLO (with labeled pedestrians, cars, traffic lights, etc.). 
(2) We endeavor to quantitatively evaluate the effect of snow on each object by introducing an indicator called snow coverage rate (SCR). In order to calculate SCR, we develop an unsupervised training strategy to train a CNN model with a distinctive activation function called $Peak \ Act$. SCR is exploited to grade the images in RSOD into four-difficulty levels (i.e., easy, normal, difficult, and particularly difficult).
Since objects covered under different degrees of snow distinctly affect the performance of detectors, we need such gradation to understand how snow degrades the accuracy of object detection.
(3) We propose a plug-and-play Cross Fusion (CF) block. Instead of relying on the traditional top-down and bottom-up operations \cite{DBLP:conf/cvpr/LinDGHHB17}, the CF block simultaneously aggregates features from different stages of the backbone. By directly fusing these features, the destroyed low-level information of the objects in the high-level features can be recovered. Besides, the CF block supports different in-out stages, making it a more flexible and adaptive plug-and-play module.
(4) We propose a lightweight object detection network named CF-YOLO by replacing the neck of YOLOv5s with CF. Experiment results on RSOD and COCO clearly show that our CF-YOLO not only has an excellent optimization ability for OD in the real-world snowy scene but also possesses an effective generalization ability.

Our main contributions are summarized as follows:
\begin{itemize}
	\item We present a real-world snowy OD dataset (RSOD). The new dataset is labeled in both the COCO and YOLO formats. To our knowledge, this is the first dataset that focuses on improving the OD accuracy in real-world snowy scenarios. 
	\item We introduce an indicator called snow coverage rate (SCR) and develop an unsupervised training strategy to train a CNN model with a distinctive activation function (Peak Act) to quantitatively evaluate the effect of snow on each object. We grade the images in RSOD into four-difficulty levels based on the evaluation. Such grading is helpful to better understand how the snow degrades the OD performance.
	\item We propose a new plug-and-play Cross Fusion (CF) block to aggregate features from different stages in a time with the flexibility of allowing different in-out stages. Users can adjust the numbers of stages, layers of CF and the size of parameters in different networks to explore the CF's performance.
	\item Many outdoor vision systems (e.g., autonomous driving, surveillance)  are required to operate smoothly in snowy weather. We propose a lightweight yet effective CF-YOLO to facilitate outdoor applications in the frequent scenarios of snow. 
\end{itemize}

The rest of this work is organized as follows. Section \ref{relatedwork} introduces the related works from four aspects: object detection in normal weather, object detection in adverse weather, image desnowing and benchmark datasets of snowy weather, and feature fusion. Section \ref{dataset} describes our constructed real-world snowy object detection dataset (RSOD). Section \ref{method} introduces the proposed CF-YOLO to promote the detection accuracy in snowy weather. Section \ref{exp} shows the sufficient experiment results to demonstrate the effectiveness of CF-YOLO. Section \ref{con} concludes this work.

\section{Related Work}
\label{relatedwork}
In this section, we first discuss the techniques of different object detectors in normal weather. Then we introduce the existing solutions for object detection in adverse conditions. After that, we discuss the existing works and datasets on snowy weather. Finally, we make a brief comparison of different feature fusion modules.

\subsection{Object Detection in Normal Weather}
Object detection predicts both the labels and bounding boxes of objects. 
Existing methods are either two-stage or one-stage based, which usually contain a backbone for feature extraction, a neck for feature fusion and a head for prediction. 
Two-stage detectors first generate regions of interest (RoIs) from the image, and then classify these RoIs by training deep networks; representative methods include R-CNN \cite{DBLP:conf/cvpr/GirshickDDM14} and its variants, such as fast R-CNN \cite{DBLP:conf/iccv/Girshick15}, faster R-CNN \cite{DBLP:journals/corr/RenHG015}, R-FCN \cite{DBLP:conf/nips/DaiLHS16}, and Libra R-CNN \cite{DBLP:conf/cvpr/PangCSFOL19}. 
Different from the highly accurate but very time-consuming two-stage detectors,
one-stage detectors utilize a single CNN to directly predict object labels and bounding boxes; representative methods include YOLO \cite{DBLP:conf/cvpr/RedmonDGF16,DBLP:journals/corr/abs-2004-10934,DBLP:yolov5}, SSD \cite{DBLP:conf/eccv/LiuAESRFB16}, RetinaNet \cite{DBLP:conf/iccv/LinGGHD17} and EfficientDet \cite{DBLP:conf/cvpr/TanPL20}. 
One-stage detectors are relatively faster with much higher FPS than two-stage detectors. DETR \cite{DBLP:conf/eccv/CarionMSUKZ20} is the first model to introduce the transformer into object detection task and regards OD as a query prediction problem.

\subsection{Object Detection in Adverse Weather}
A detector trained on clean images usually fails to yield desirable results under adverse weather conditions (e.g., snowy, rainy, hazy and low-light), due to the domain shift in input images \cite{eccv/SindagiOYP20}. Currently, there are mainly three solutions to alleviate image degradation.
The first solution is to dilute the effect of weather-specific information by a pre-processing step, such as image desnowing/derainig/dehazing \cite{DBLP:journals/corr/abs-2103-11298,DBLP:journals/tcsv/JawHK21} or low-light image enhancement \cite{DBLP:journals/corr/abs-2112-06451}.
As known, both the weather-specific information and image details (an important cue for object detection) are both of high frequency and small in scale.
Although complicated image restoration models are designed and trained on synthetic data with strong pixel-level supervision, they still lose image details easily. 
The second solution is to jointly learn image restoration and object detection by two-branch networks \cite{HuangLJ21}, where the two branches share the feature extraction layers. However, it is hard to balance the two tasks during training.
The third solution is to exploit unsupervised domain adaptation \cite{Hnewa21} to align the features of clean images (sources) and images captured under adverse weather (targets). However, the latent information that is beneficial for detection is often ignored during image restoration.  

\subsection{Snowy Datasets}
Due to the lack of real-world paired snow/snow-free images, existing datasets on snowy weather are generally obtained by adding snow masks to clean images, 
like Snow100K in \cite{DBLP:journals/tip/LiuJHH18}, SnowKITTI2012 \cite{DBLP:journals/corr/abs-2103-11298}, and SnowCityScapes \cite{DBLP:journals/corr/abs-2103-11298}. 
Existing snow removal methods trained on these synthetic datasets always have good performance but deteriorate significantly on real-world snowy images. Meanwhile, they only remove snow in the air, ignoring the fact that objects are often covered by snow which affects the performance of object detection.
Moreover, the results of these methods will lose image details (an important cue for high-level vision tasks), since image details and snow are both of high frequency and small in scale. Therefore, they inadequately improve the performance of down-stream applications. 

\subsection{Feature Fusion}
Existing feature fusion works contain FPN \cite{DBLP:conf/cvpr/LinDGHHB17}, PANet \cite{DBLP:conf/cvpr/LiuQQSJ18}, NAS-FPN \cite{DBLP:conf/cvpr/GhiasiLL19}, BiFPN \cite{DBLP:conf/cvpr/TanPL20}, ASFF \cite{DBLP:journals/corr/abs-1911-09516}, etc. 
FPN integrates features from different stages of the backbone through a top-down path. Based on FPN, PANet exploits a bottom-up path augmentation to enhance the entire feature hierarchy. The BiFPN Layer \cite{DBLP:conf/cvpr/TanPL20} is developed for easy and fast multi-scale feature fusion by bidirectional cross-scale connections. OctConv \cite{DBLP:conf/iccv/Chen0XYKRYF19} decomposes features into different spatial frequencies to improve the efficiency of CNNs. gOctConv \cite{DBLP:conf/eccv/GaoTCLCY20} is proposed which possesses the advantage of flexible feature fusion for the arbitrary in-and-out branches.  In this work, we exploit gOctConv as the fundamental component of our cross fusion block, which exhibits a good ability for feature fusion.

\section{Real-world Snowy Object Detection Dataset}
\label{dataset}
\subsection{Dataset Introduction}
The established real-world snow object detection dataset, called RSOD, contains 2100 images captured in various real-world snowy scenes.
To make RSOD convenient to use for public studies, the labels are fully compatible with MSCOCO and we provide both COCO and YOLO formats. Fig. \ref{dataset} shows the label distributions in RSOD. Note that most of our captured snowy images are about townscape and traffic scenes.

\begin{figure}[ht] \centering
	\includegraphics[width=1\linewidth]{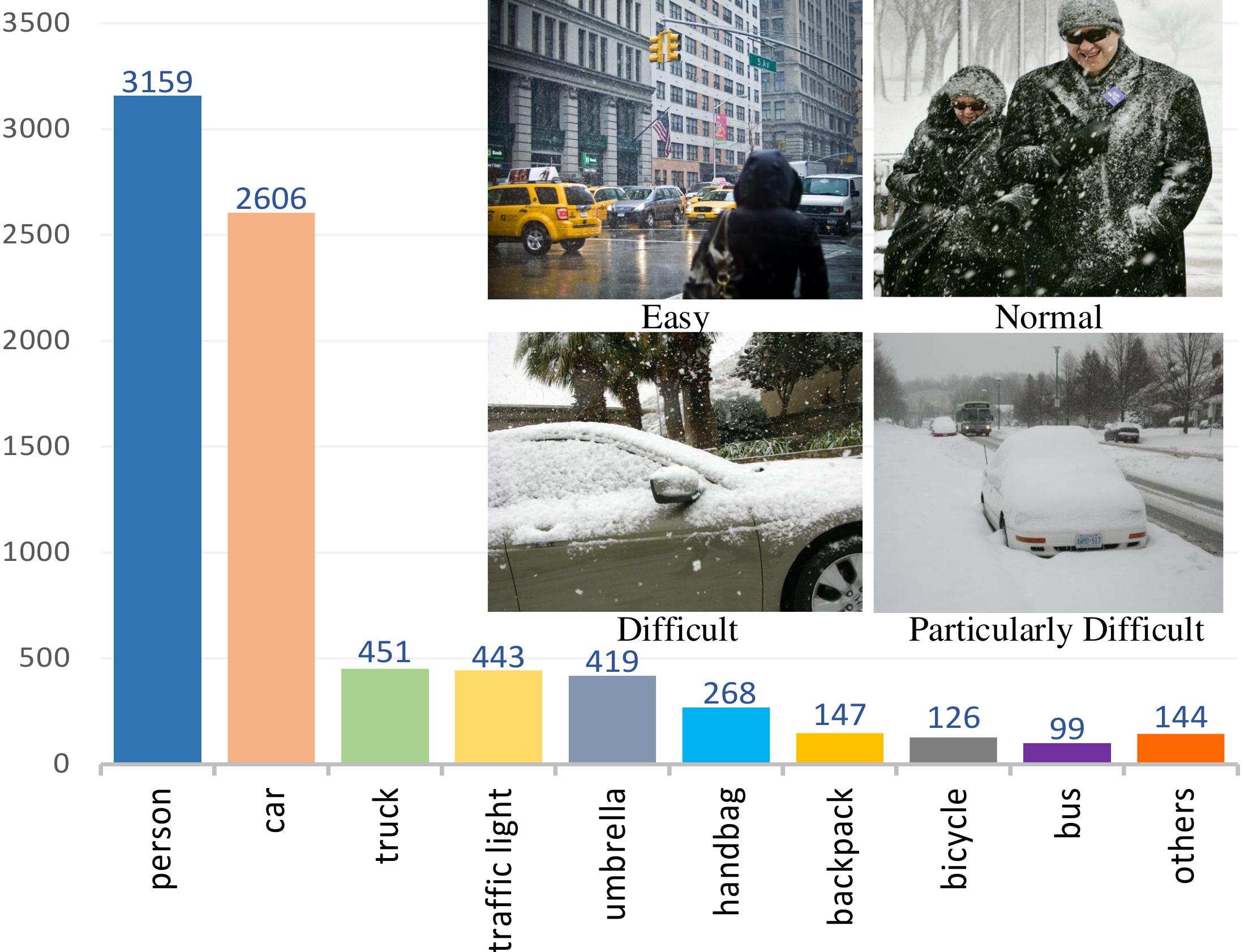}
	\caption{
	The label distributions in the proposed RSOD. Most of the annotated objects are cars (2606 labeled), pedestrians (3159 labeled) and traffic lights (443 labeled). The detectors perform differently under different snow coverage rates. We divide snow images into four difficulty levels (i.e. easy, normal, hard, and particula hard) according to the snow coverage rates.}
	\label{dataset1}
\end{figure}

Considering the fact that the performance of cutting-edge detectors will deteriorate when objects are covered under different degrees of snow,
we grade the snowy images into four difficulty levels, i.e., easy, normal, difficult, and particularly difficult. Such gradation facilitates determining the effect of snow on objects, which helps to study how snow degrades the accuracy of object detection.

How to grade the snowy images objectively is the main challenge to construct the dataset.
In order to conduct a fair gradation of the difficulty levels, we introduce an indicator called SCR, which is used to calculate the snow coverage rate (SCR) of different objects.
By combining SCR and human observation, we grade image numbers 1\textasciitilde600 to the easy level, 601\textasciitilde1600 to the normal level, 1601\textasciitilde2000 to the difficult level and 2001\textasciitilde2100 to the particularly difficult level. Fig. \ref{dataset1} shows the typical images of different levels.

\begin{figure*}[htb!]
\centering
\subfigure[Input]{
\includegraphics[width=4.8cm]{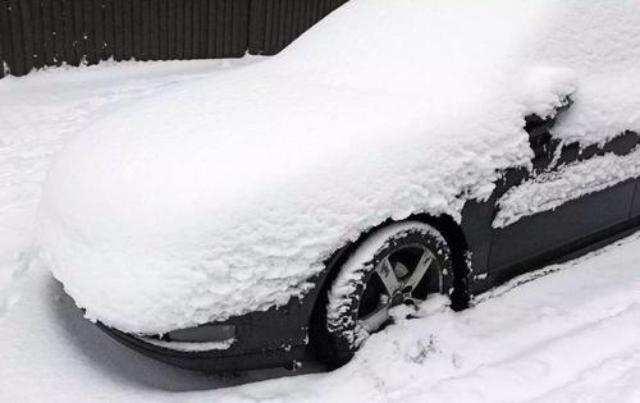}
}
\subfigure[CH-31 and SCR]{
\includegraphics[width=4.8cm]{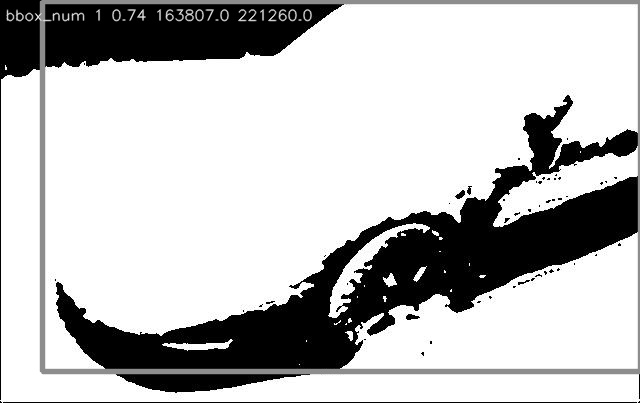}
}
\subfigure[CH-11 and SCR]{
\includegraphics[width=4.8cm]{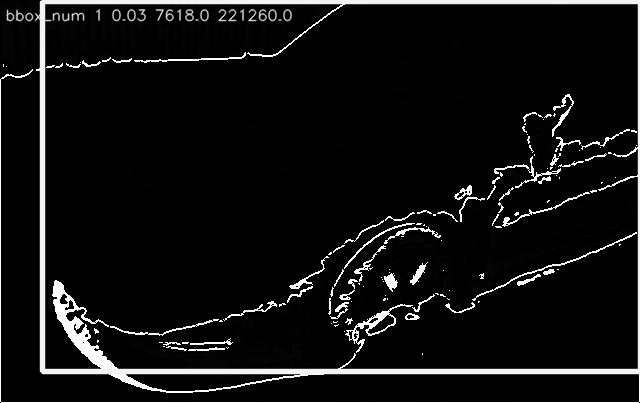}
}
\subfigure[Peak Act]{
\includegraphics[width=4.9cm]{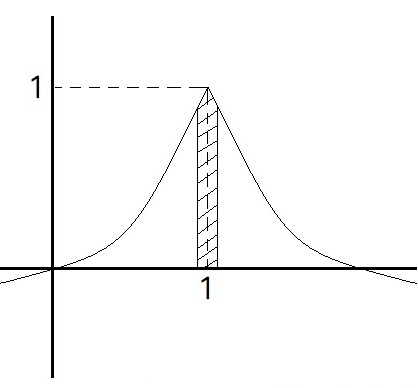}
}
\subfigure[CH-31 3D]{
\includegraphics[width=4.9cm]{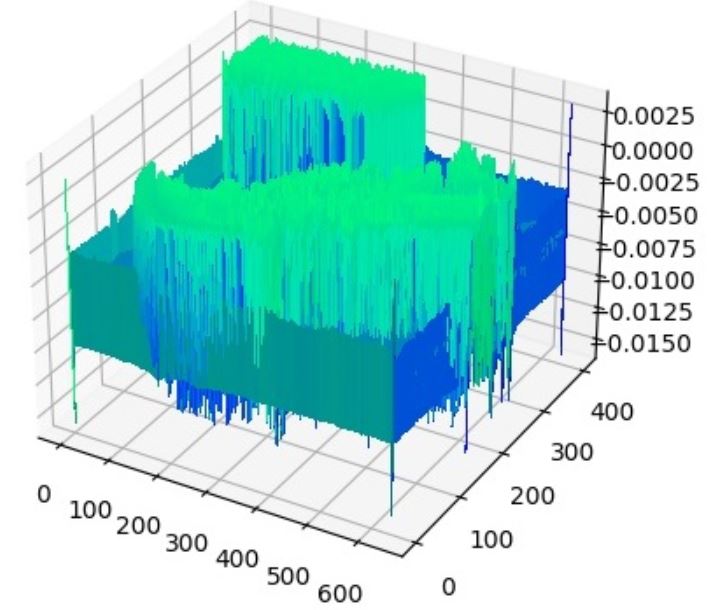}
}
\subfigure[CH-11 3D]{
\includegraphics[width=4.9cm]{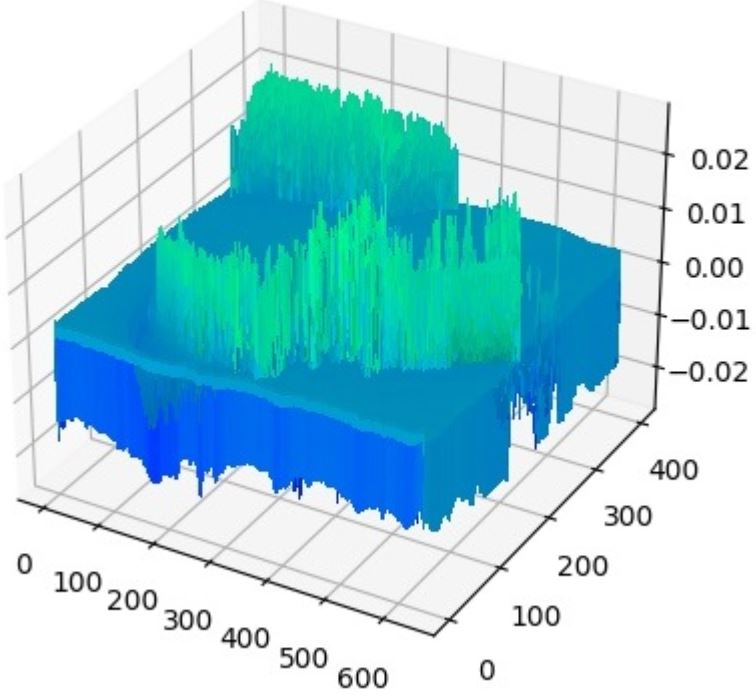}
}
\caption{The visualized results of the CNN model. Different channels respond to different image features. The channel-31 (CH-31) responds to snow exclusively and provides the snow coverage rate of the object. The channel-11 (CH-11) responds to the edge of the object.}
\label{peak_act}
\end{figure*}

\subsection{Unsupervised Training for SCR Calculation}
Quantitatively evaluating the effect of snow on the covered objects is challenging even for humans, since there are no benchmarks for the evaluation. We assume that the benchmark of snow effect depends on the snow coverage rate (SCR) in the object's bounding box. Therefore, SCR can be formulated as $SCR = A_{snow} / A_{bbox}$, where $A_{snow}$, $A_{bbox}$ represent the areas of snow and bounding box, respectively. 

Calculating SCR directly is challenging because it costs a lot to label out all snow in the dataset.
To address this problem, we develop an unsupervised training strategy to train a CNN which responds to snow pixels and depresses non-snow pixels. Inspired by sparse coding, our strategy contains three essential factors. 

\textbf{First}, to respond to snow pixels, we train a CNN model with images that heavy snow covers most of the image area, where the corresponding ground truth is a map with the same size of the input image and with all pixels being equal to 1. This step will guide the model to map each pixel to the value of 1, and the convolution kernels in the model will encode snow features through back propagation. 

\textbf{Second}, to depress non-snow pixels, we design an activation function with a very narrow activation bandwidth, which allows the response of the convolution kernels to some specific features like snow, and depress other features. As shown in Fig. \ref{peak_act}(d), we call the activation function $Peak \ Act$. Since snow covers the largest area of the images we use to train, it is natural that convolution kernels will respond to snow pixels and depress non-snow pixels. 

The function of Peak Act lies on the following three rules: 
\begin{itemize}
	\item Be a peak function where the peak is $(1, 1)$. Because our ground truth is a matrix with all elements being equal to 1, the training process will guide the outputs towards 1. And the peak will constrain the effective area in a very small bandwidth, as shown in Fig. \ref{peak_act}(d).
	\item Zero maps to zero. If a zero mapping to a non-zero value, there will be some $lazy$ convolution kernels with all weights being equal to 0, which smooth all the pixels to a non-zero value. Therefore, features of later layers can be easily equal to the ground truth, which leads to the failure of the training.
	\item Be a concave function to make sure that feature values will not get closer to 1 after passing through the activation function. Features can only get closer to 1 through optimization.
\end{itemize}

The proposed Peak Act is defined as:
\begin{eqnarray}
f(x) =\begin{cases}
0.2x & x < 0 \\
x^2 & 0 \leq x < 1 \\
(x-2)^2 & 1 \leq x < 2 \\
-0.2(x-2) & x \geq 2
\end{cases}
\end{eqnarray}
Different from many other activation functions, Peak Act is not a monotone increasing or a continuously differentiable function. 
Our motivation is simple: by limiting the activation bandwidth of the convolution kernel weights in a very narrow space, we can obtain a sparse and finite weight matrix that responds to snow flexibly. 

\textbf{Third}, the last layer of the CNN is a $Max-out$ function, which will output the maximal feature value of every pixel in the channel dimension, and form a one-channel feature map $O$ to compute the loss with the ground-truth ($GT$).
Due to the fact that the upper limit of Peak Act is 1, the output of the network will be always less than or equal to 1. The $Max-out$ layer will encourage different channels to respond to different features, leading to highly specific optimizations of the kernels. 
We define the loss as:
\begin{equation}
    Loss = \alpha\frac{1}{W*H}\sum_{i}\sum_{j} (GT_{i j} - O_{i j}) + \beta\Vert P \Vert_1
\end{equation}
where $P$ denotes the network's parameters, $\alpha=1$, and $\beta=0.0001$ are the weights to balance the two terms. 
The first term is to guide the optimization direction, and the second is an $L_1$ regularization to get a sparse feature. 

\begin{figure*}[ht]
	\centering
	\includegraphics[width=\linewidth]{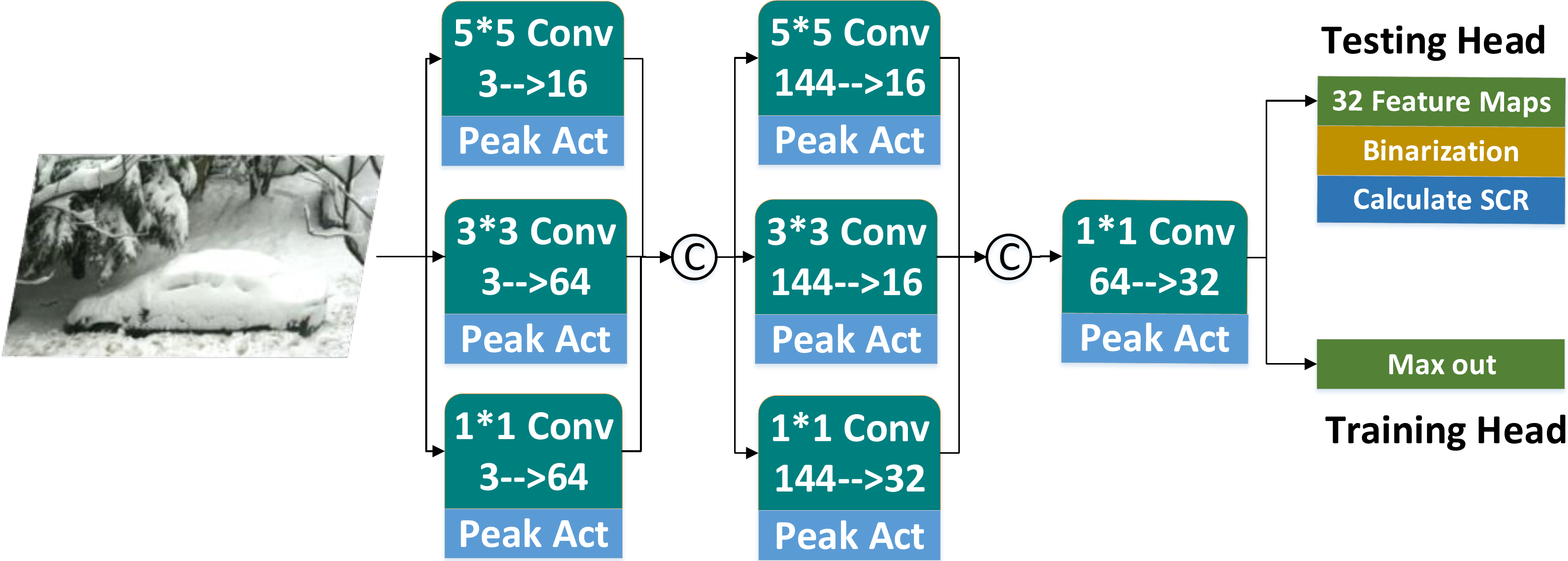}
	\caption{The structure of CNN in our unsupervised training strategy. During training, we use the output of the training head, a $Max_Out$ layer, to compute the loss. While testing (i.e., calculating SCR), we use the output of the testing head.}
	\label{snow_rate}
\end{figure*}
The CNN model is shown in Fig. \ref{snow_rate}. During training and testing (i.e., calculating SCR), we exploit different heads. The input image is decomposed into 32 channels by the model, and we binarize and visualize the feature maps of different channels. As shown in Fig. \ref{peak_act}(b) and Fig. \ref{peak_act}(c), the Feature Map-31 responds to snow very specifically, while the Feature Map-11 responds to edges. The feature maps which respond to snow can be used to calculate SCR by counting the light pixels in the binarized maps. We also visualize the 3D surfaces of different channels, as shown in Fig. \ref{peak_act}(e) and Fig. \ref{peak_act}(f), it clearly shows that the channel-31 responds to the snow area and depresses the non-snow area distinctively. 

The utilization of Peak Act and the proposed CNN model with an unsupervised training strategy are essential to calculate the SCR and grade the snowy images. More study will be addressed in Section \ref{exp}.
\section{Methodology}
\label{method}
Recently, cutting-edge detectors achieve remarkable progress and strengthen many outdoor vision systems, e.g., autonomous driving, and surveillance. But these methods also suffer from various adverse weather and fail to yield desirable results. Since models trained under normal circumstances do not assure an advantage over the contaminated scene. As shown in Tab. \ref{tab1}, the performance of various detectors trained on MSCOCO degrades significantly on RSOD, due to the domain shift problem. 

\begin{figure*}[ht]
	\centering
	\includegraphics[width=\linewidth]{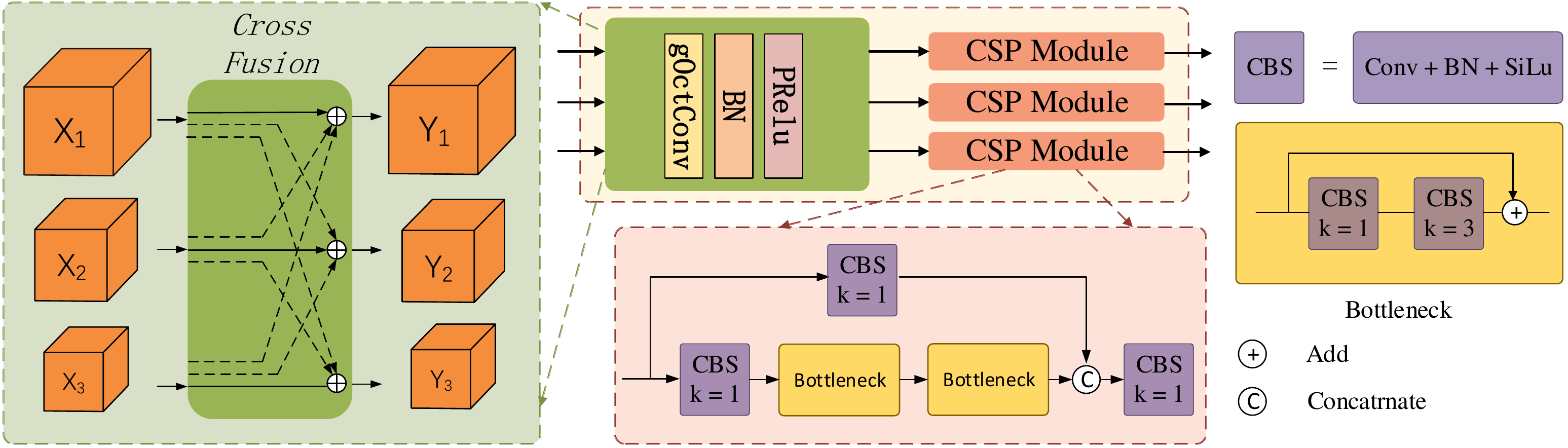}
	\caption{The structure of Cross Fusion. $X_1, X_2, X_3, Y_1, Y_2, Y_3$ indicate the different stages of input and output branches. The features from each level are fed into a gOctConv (without sharing weights), a Batch Normalization layer, and a PRelu activation function. The post-processing component of the CF block is Cross Stage Partial (CSP)~\cite{DBLP:conf/cvpr/WangLWCHY20}.}
\label{model}
\end{figure*}

Put the enormous degradation aside, we also find that some large objects are more likely to be omitted by YOLOv5s in snowy images. This may violate our common sense because many studies have shown that detection models perform better on large objects.
Based on this observation, we make a small adjustment to YOLOv5s. By setting the detection confidence threshold to 0.01, we surprisingly find that YOLOv5s has already detected those large objects in the snowy images (similar phenomena also exist in many other snowy images), but the confidence is too low to pass Non-Maximum Suppression (NMS), leading to the mis-prediction, see Fig. \ref{pre_ana}(b) for instance. The reason is that heavy snow can change the outline, texture and surface of objects, missing and distorting the low-level vision information.

Yolov5s uses the Feature Pyramid and Path Aggregation Network (FPN+PANet) as the feature fusion module. Features pass through a top-down and bottom-up route before the YOLO prediction head. According to the structure of YOLOv5s, objects of different sizes are predicted in different stages, and large objects are predicted in the last stage, which means large object features pass through the deepest network. In snowy images, low-level vision information is missed and distorted, and this meaningless information will propagate along with the network. In a deep network, the receptive field of deep layers is relatively large, therefore, deeper layers are likely to take more meaningless features into account. This may $dilute$ meaningful features, interfere with the network to extract meaningful features from objects covered by snow, and decrease the confidence of the prediction.

\subsection{Cross Fusion}
To address this problem, we propose a new Cross Fusion (CF) block that can directly integrate the features from different levels. The purpose of this module is to shorten the propagation route instead of making the model more complicated or deeper. This alleviates the $dilution$ of meaningful features when the network goes deeper. In detail, inspired by \cite{DBLP:conf/eccv/GaoTCLCY20}, we utilize gOctConv as the fusing component of the CF module.
As shown in Fig. \ref{model}, the input feature maps with different scales are simultaneously fed to the CF layer, encouraging the last stage to reach low-level features directly. The CF layer also allows different in-out branches, making CF a flexible plug-and-play module to adapt to different models. The post-processing component is the Cross Stage Partial (CSP) module. 

Compared with the top-down and bottom-up structure of ``FPN+PANet", CF can provide a shorter route between low-level and high-level features. One of the feature fusion processes of CF can be expressed as:
\begin{equation}
\begin{aligned}
    O_1 &=f_{CSP}(Conv_{11}(X_1)\oplus Conv_{12}(Resize(X_2)) \\
            &\oplus Conv_{13}(Resize(X_3)))
\end{aligned}
\end{equation}
$f_{CSP}$ denotes the CSP module and $\oplus$ means the element-wise addition. $O_1$ is the up-branch of CF outputs, and other branches have the same expression. Feature fusion of CF happens before the post-processing component, while feature fusion of ``FPN+PANet"  can only happen sequentially along with the top-down and bottom-up operations.
\begin{figure}[h!]
	\centering
	\includegraphics[width=\linewidth]{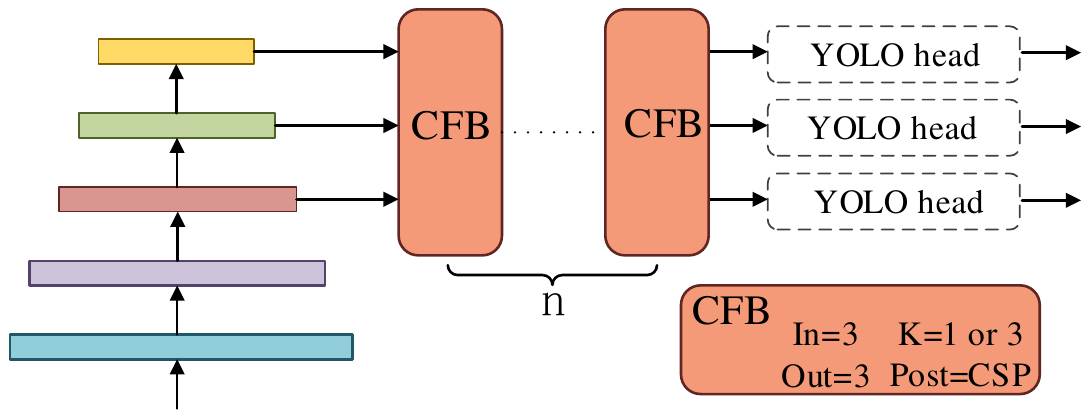}
	\caption{The architecture of the proposed CF-YOLO. Modifying the number of CF (n), in-out stages (In and Out), and the kernel sizes (K) to obtain different versions of CF-YOLO for different usages.}
	\label{CF-YOLO}
\end{figure}


\subsection{CF-YOLO}
We replace the neck of YOLOv5s with CF and propose CF-YOLO (see Fig. \ref{CF-YOLO}). Besides, the structure of CF-YOLO is very flexible. It can be easily modified by changing the number of CF (n), in-out stages (In and Out), and the kernel size of gOctConv (K). In this work, our CF-YOLO stacks two layers of CF (n=2). CF-YOLO (K=1), CF-YOLO (K=3) represents that the kernel sizes of CF equals to 1 and 3 respectively.

\section{Experiments and discussion}
\label{exp}
\subsection{Comparison of different activation functions}
Since the input images to the network used to calculate SCR are heavy snow images, the statistical distribution of training data concentrates on the feature of snow, which encourages the network to respond to snow. 
To verify the effectiveness of our proposed Peak Act, we compare the general activation functions, namely Sigmoid, ReLU \cite{2011Deep} and Leaky ReLU \cite{maas2013rectifier} with Peak Act. And we select the best feature map visualization results for comparison.
\begin{figure*}[htb]
\centering
\subfigure[Best feature map using Sigmoid]{
\includegraphics[width=3cm]{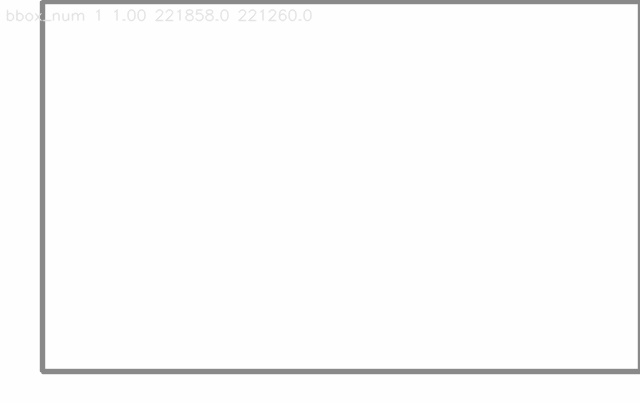}
}
\subfigure[Best feature map using ReLU]{
\includegraphics[width=3cm]{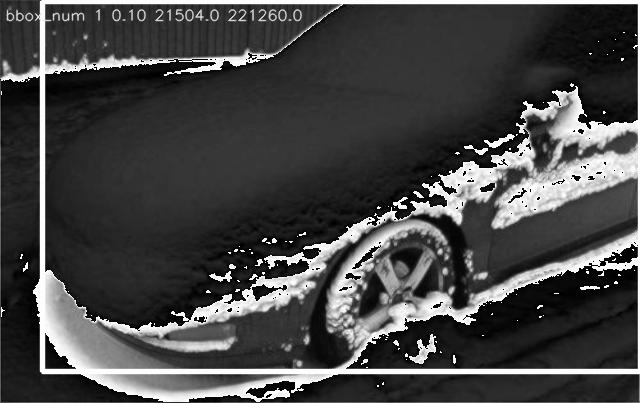}
}
\subfigure[Best feature map using Leaky ReLU]{
\includegraphics[width=3cm]{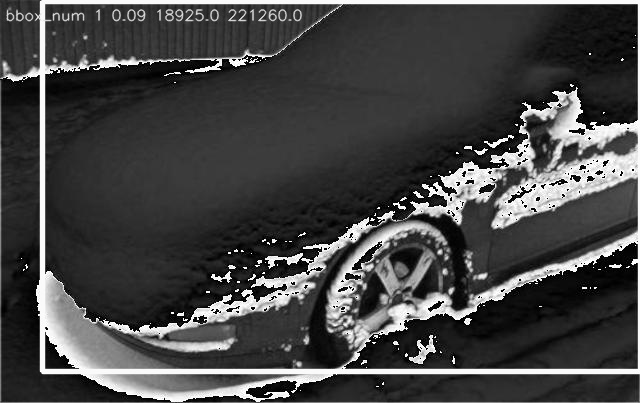}
}
\subfigure[Best feature map using Leaky ReLU and Peak Act in last layer]{
\includegraphics[width=3cm]{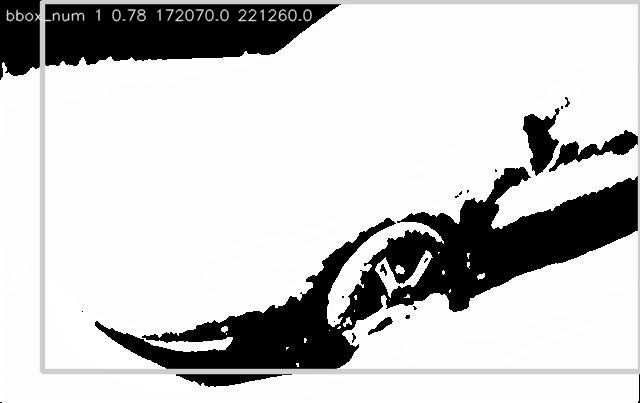}
}
\subfigure[Best feature map using Peak Act]{
\includegraphics[width=3cm]{featureX31.jpg}
}

\subfigure[3D surface of best channel using Sigmoid]{
\includegraphics[width=3cm]{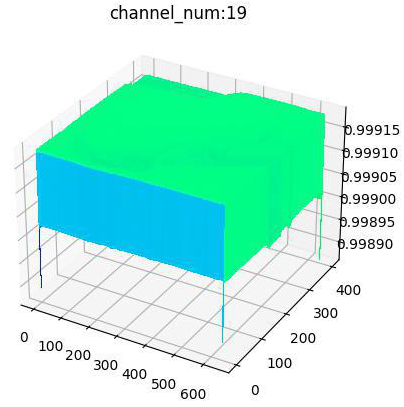}
}
\subfigure[3D surface of best channel using ReLU]{
\includegraphics[width=3cm]{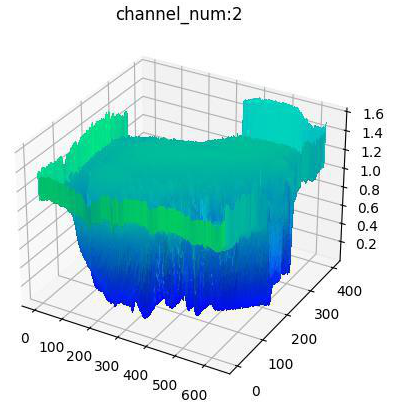}
}
\subfigure[3D surface of best channel using Leaky ReLU]{
\includegraphics[width=3cm]{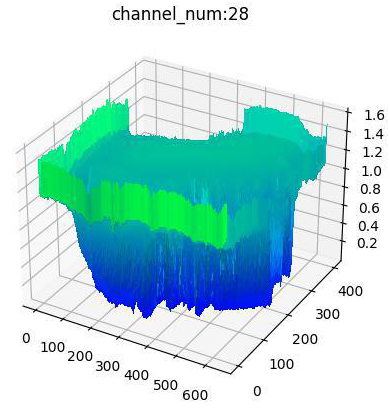}
}
\subfigure[3D surface of best channel using Leaky ReLU and Peak Act in last layer]{
\includegraphics[width=3cm]{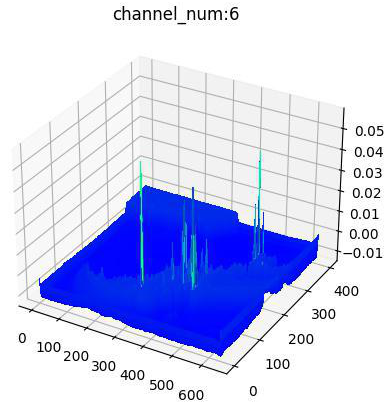}
}
\subfigure[3D surface of best channel using Peak Act]{
\includegraphics[width=3cm]{channel_num_31.jpg}
}
\caption{The visualized results of the CNN model with different activation functions. The model with Peak Act generates the clearest feature map to calculate the SCR.}
\label{act_cmp}
\end{figure*}

As shown in Fig. \ref{act_cmp}, the output of the CNN models using different activation functions performs distinctively. Sigmoid (column 1) cannot identify the difference between snow and non-snow areas. The best feature map almost equals to $GT$, which is a matrix with all elements equal to 1.
For ReLU (column 2) and Leaky ReLU (column 3), feature values of different areas are not distinctive enough, thus we cannot calculate the SCR.
The model with Leaky ReLU in former layers and Peak Act in the last layer (column 4) shows a much better performance than networks that only use general activation functions. But the details of the best feature map are not as good as the network using Peak Act only. Only Peak Act can ensure that the model separates out the snow features from non-snow features. We have to raise this question: what makes Peak Act unique?

\begin{figure*}[h]
\centering
\subfigure[Peak Act]{
\includegraphics[width=4cm]{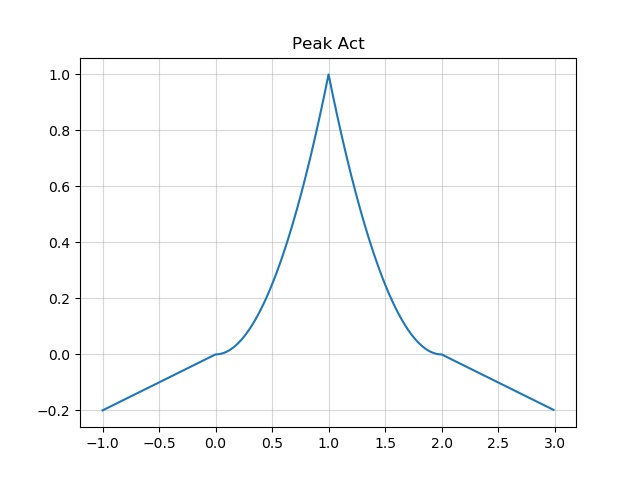}
}
\subfigure[Sigmoid]{
\includegraphics[width=4cm]{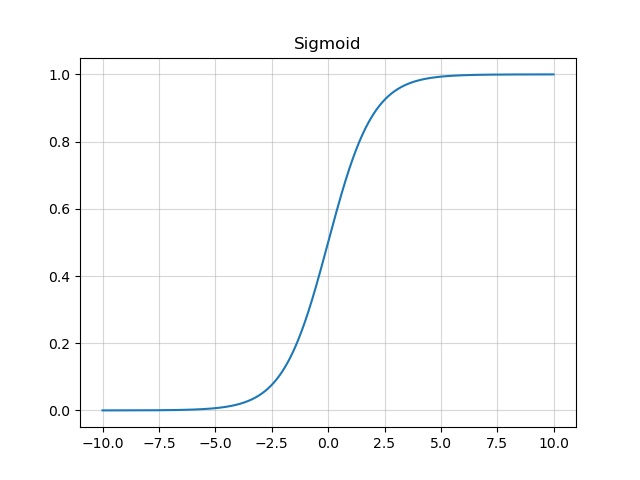}
}
\subfigure[ReLU]{
\includegraphics[width=4cm]{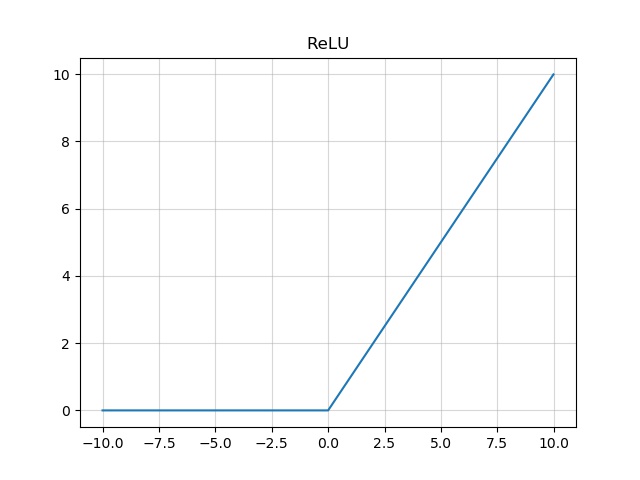}
}
\subfigure[Leaky ReLU]{
\includegraphics[width=4cm]{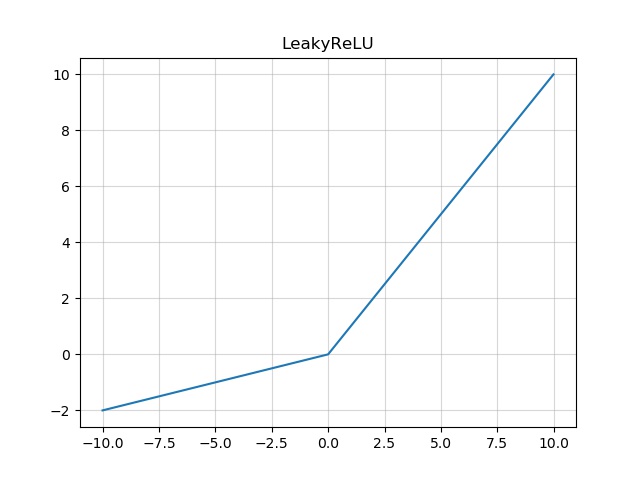}
}
\subfigure[Peak Act gradient]{
\includegraphics[width=4cm]{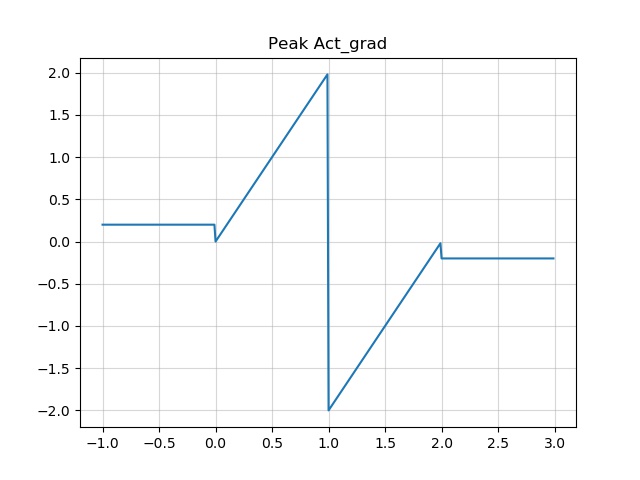}
}
\subfigure[Sigmoid gradient]{
\includegraphics[width=4cm]{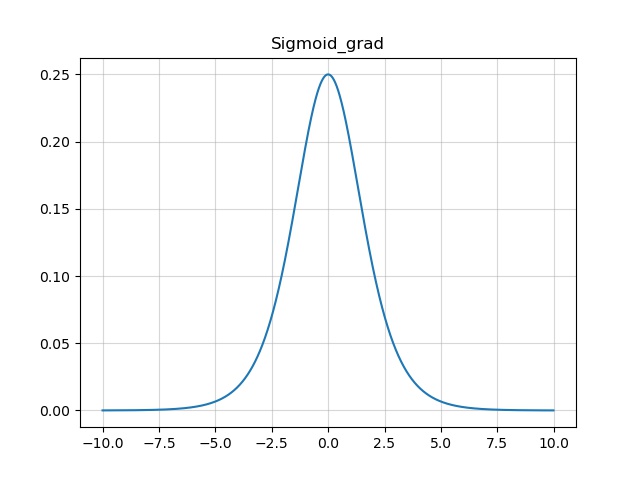}
}
\subfigure[ReLU gradient]{
\includegraphics[width=4cm]{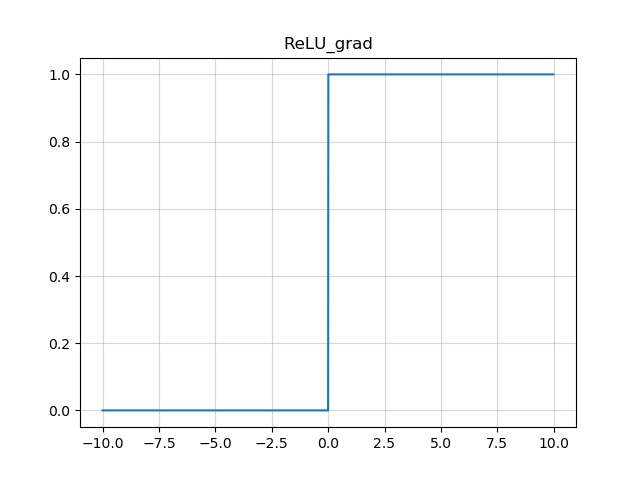}
}
\subfigure[Leaky ReLU gradient]{
\includegraphics[width=4cm]{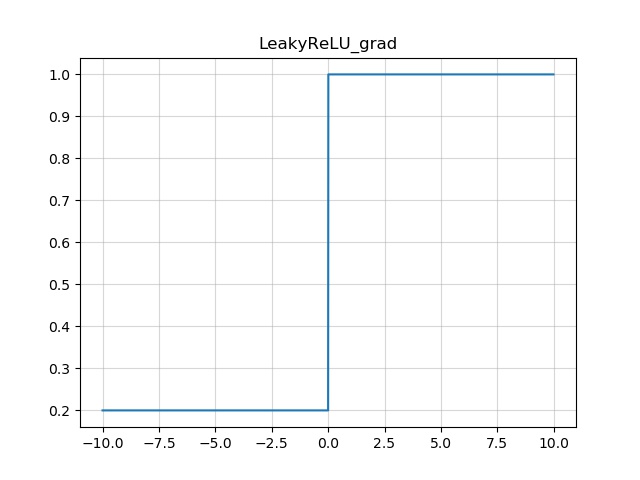}
}
\caption{Different activation functions and their gradients.}
\label{act_func}
\end{figure*}

As shown in Fig. \ref{act_func}, Peak Act has the following characters compared with the general activation functions:
\begin{itemize}
	\item Peak Act has no vanishing gradient problem. The function has a non-zero and great derivative in the whole domain. Making the CNN we proposed in the unsupervised training strategy easy to train.
	\item The range of Peak Act is [$-\infty$, 1]. It is a finite function towards the training direction. This helps to maintain the value of weights great. In the CNN, the feature value is always less than 1, leading the network to keep the value of weights great in order to keep the output close to 1. And greater weight value enlarges the differences between respond areas and non-respond areas.
	\item The ground truth we generate in the unsupervised training strategy is the peak of Peak Act. This limits the output of convolutions in a very narrow bandwidth, which leads to the sparsity of kernels. With Peak Act, different channels have to concentrate on specific features.
\end{itemize}

Sigmoid is a finite function towards the training direction like Peak Act. Since the $\lim\limits_{N\to+\infty} = 1$, this will lead the vanishing gradient problem. Moreover, it allows `lazy' kernels to have very great weight values, forcing the outputs of every pixel close to 1. And there will be no differences between snow areas and non-snow areas.

For Leaky ReLU, it has no limit or vanishing gradient problem, but it is an infinite function towards the training direction. 
When the feature value is very large, the weight value will be small, which will narrow the gap between snow and non-snow areas.

For ReLU, it has the disadvantages of both Sigmoid and Leaky ReLU.

\begin{figure*}[h]
\centering
\subfigure[]{
\includegraphics[width=4cm]{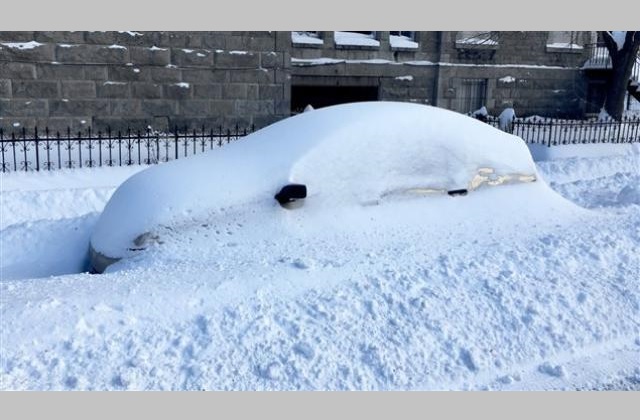}
}
\subfigure[]{
\includegraphics[width=4cm]{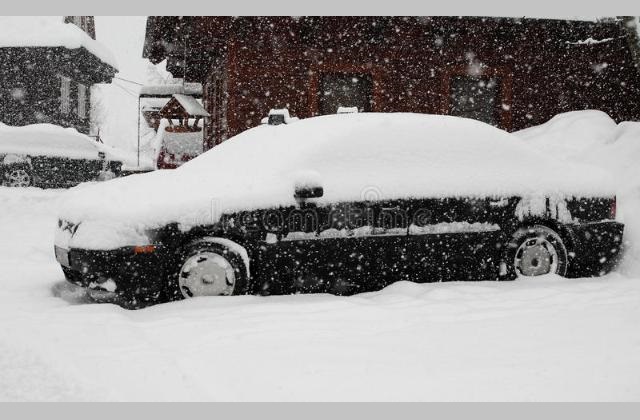}
}
\subfigure[]{
\includegraphics[width=4cm]{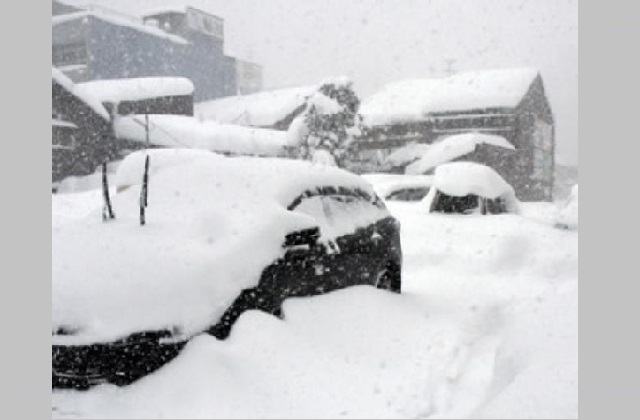}
}
\subfigure[]{
\includegraphics[width=4cm]{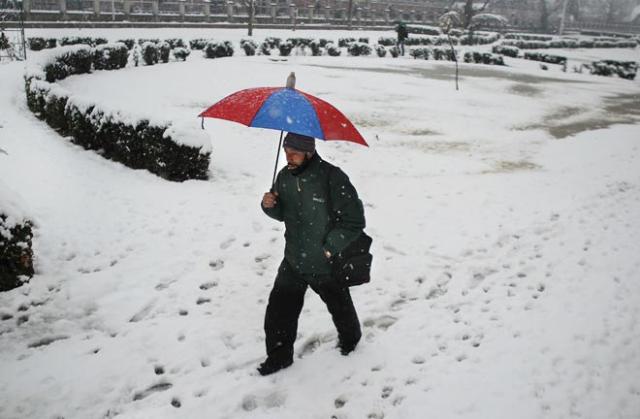}
}
\subfigure[]{
\includegraphics[width=4cm]{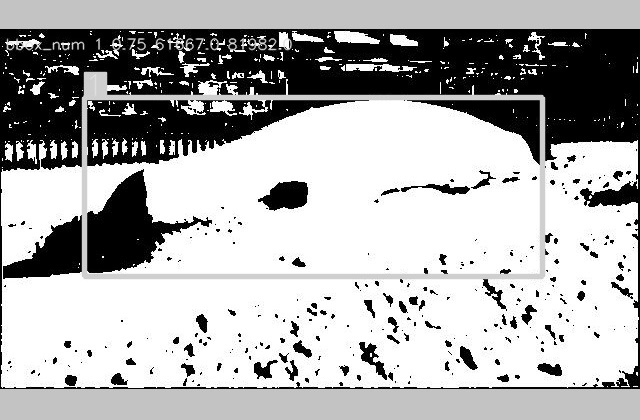}
}
\subfigure[]{
\includegraphics[width=4cm]{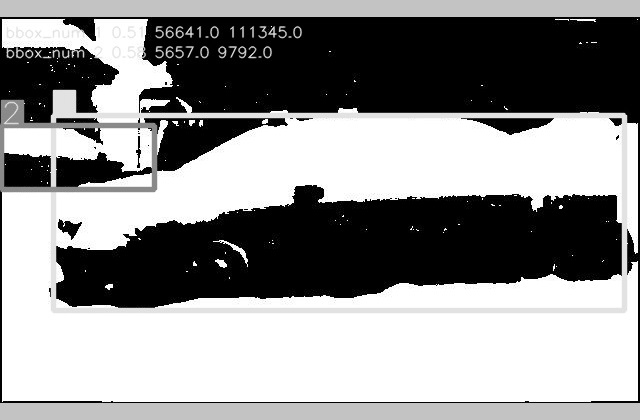}
}
\subfigure[]{
\includegraphics[width=4cm]{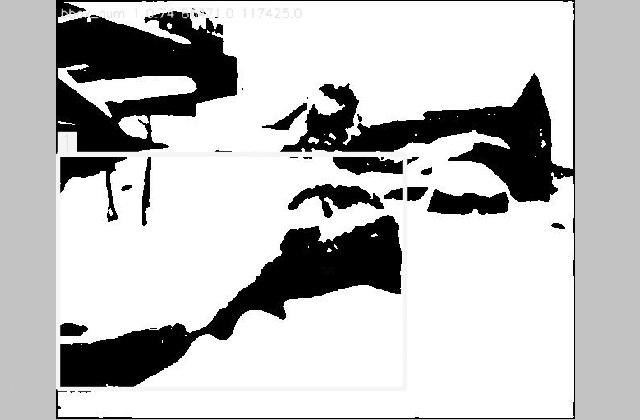}
}
\subfigure[]{
\includegraphics[width=4cm]{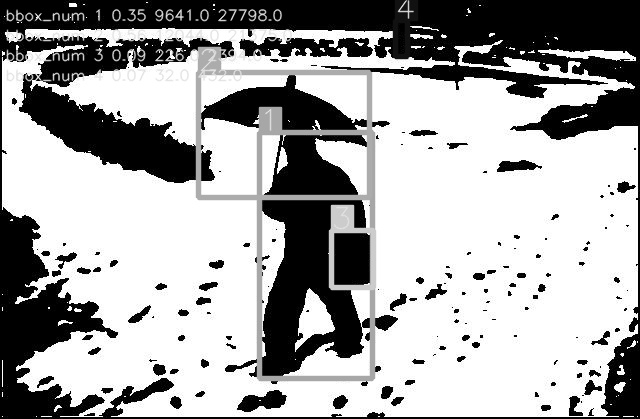}
}
\caption{The visualized results of the proposed model with the unsupervised training strategy. SCR is marked in the generated maps.}
\label{SCR}
\end{figure*}
All of the above reasons make the proposed Peak Act unique. By leveraging the Peak Act and the unsupervised training strategy, we obtain the maps using only dozens of snow images and a few minutes of training time, making quantitatively evaluating snowy images possible.
Our network can figure out the snow areas in images that are not used during training. As shown in Fig. \ref{SCR}, the network successfully generates feature maps of snow and enables the calculation of SCR. For image (a), the $SCR_{1}=0.75$, here $SCR_{1}$ indicates the SCR of object 1. For image (b), the $SCR_{1}=0.51$, $SCR_{2}=0.58$. For image (c), the $SCR_{1}=0.74$. For image (d), the $SCR_{1}=0.35$, $SCR_{2}=0.56$, $SCR_{3}=0.09$, $SCR_{4}=0.07$. Image (a) and (c) get much higher SCR, which coincides with human observation. It proves that our unsupervised training strategy can quantitatively evaluate the effect of snow.

\begin{table*}[ht]
\caption{Comparison with existing methods for object detection on our proposed dataset. Detectors (pre-trained on MSCOCO) show very different performances on RSOD. Our dataset with the four-difficulty levels provides a quantitative way to study the degradation of detectors in real-world snow. YOLO-based Detectors averagely work better.}
\centering
\resizebox{1.8\columnwidth}{!}{
\begin{tabular}{c|c c|c c|c c|c c|c c}
    \Xhline{1.2pt}
        \multirow{2}{*}{Method} &\multicolumn{2}{c|}{Easy (600)} &\multicolumn{2}{c|}{Normal (1000)} &\multicolumn{2}{c|}{Difficult (400)} &\multicolumn{2}{c|}{Particularly  Difficult (100)} &\multicolumn{2}{c}{All Levels (2100)} \\ 
        \cline{2-11}
            & $AP$ & $AP_{50}$ & $AP$ & $AP_{50}$ & $AP$ & $AP_{50}$ & $AP$ & $AP_{50}$ & $AP$ & $AP_{50}$\\ 
    \hline
        YOLOv5s &41.1 &61.3 &\textbf{36.1} &\textbf{55.5} &\textbf{26.4} &41.3 &25.2 &39.2&\textbf{34.3}&\textbf{52.0} \\ 
        EfficientDet D0 (CVPR'20) &20.1 &28.6 &22.9 &34.4 &17.1 &27.1 &29.1&41.0 &18.5 &27.2 \\
        EfficientDet D1 (CVPR'20) &22.4 &30.1 &26.6 &37.6 &18.7 &29.5 &26.6 &\textbf{45.5} &21.1 &29.9 \\
        SSD300 (ECCV'16) &26.1&44.2 &27.1&46.8 &16.8 &29.8 &20.5&37.4 &23.1 &40.2 \\  
    \hline              
        CF-YOLO (K=1) &\textbf{45.6} &\textbf{67.5} & 34.9&55.0& \textbf{27.7} &\textbf{44.4} &\textbf{30.7} &\textbf{47.1} &\textbf{34.5}&\textbf{53.4} \\ 
        CF-YOLO (K=3) &\textbf{63.8}&\textbf{72.3} &\textbf{35.5} &\textbf{56.0}& 26.3&\textbf{41.5} &\textbf{31.9}&44.2 &32.7&50.2 \\
    \Xhline{1.2pt}
\end{tabular}} \label{tab1}
\end{table*}

\begin{figure*}[ht]
	\centering
	\includegraphics[width=\linewidth]{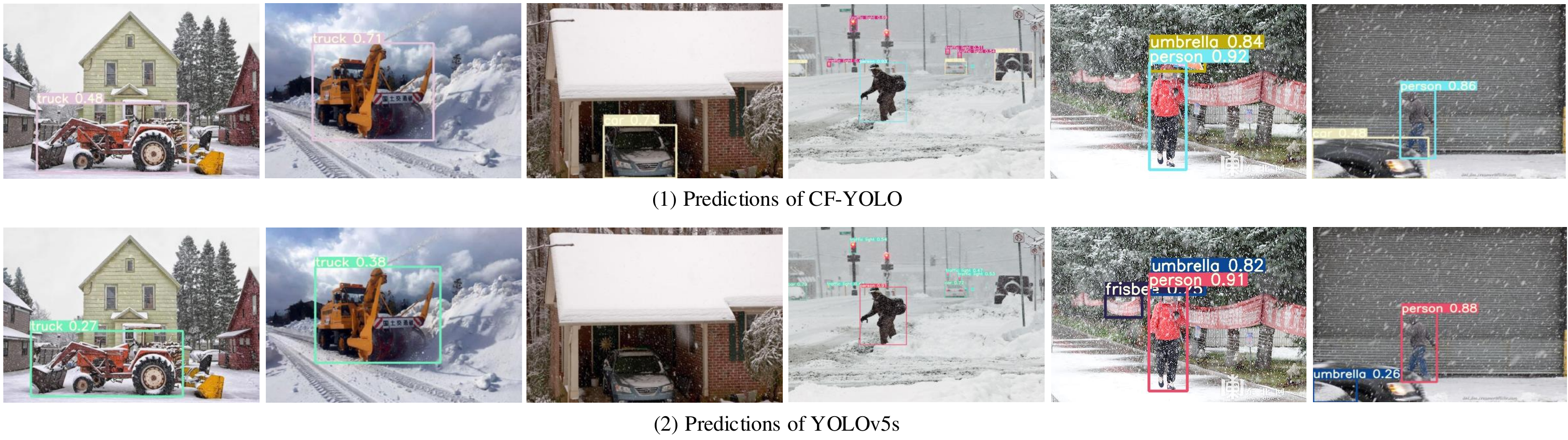}
	\caption{The predicted results of CF-YOLO and YOLOv5s trained on MSCOCO only. Compared with YOLOv5s, our method: (1) has higher confidence in detection results (columns 1 and 2); (2) can reduce missed detections (columns 3 and 4); (3) can reduce false detections (columns 5 and 6).}
	\label{prediction}
\end{figure*}

\begin{figure}[htb]
\centering
\includegraphics[width=1\columnwidth]{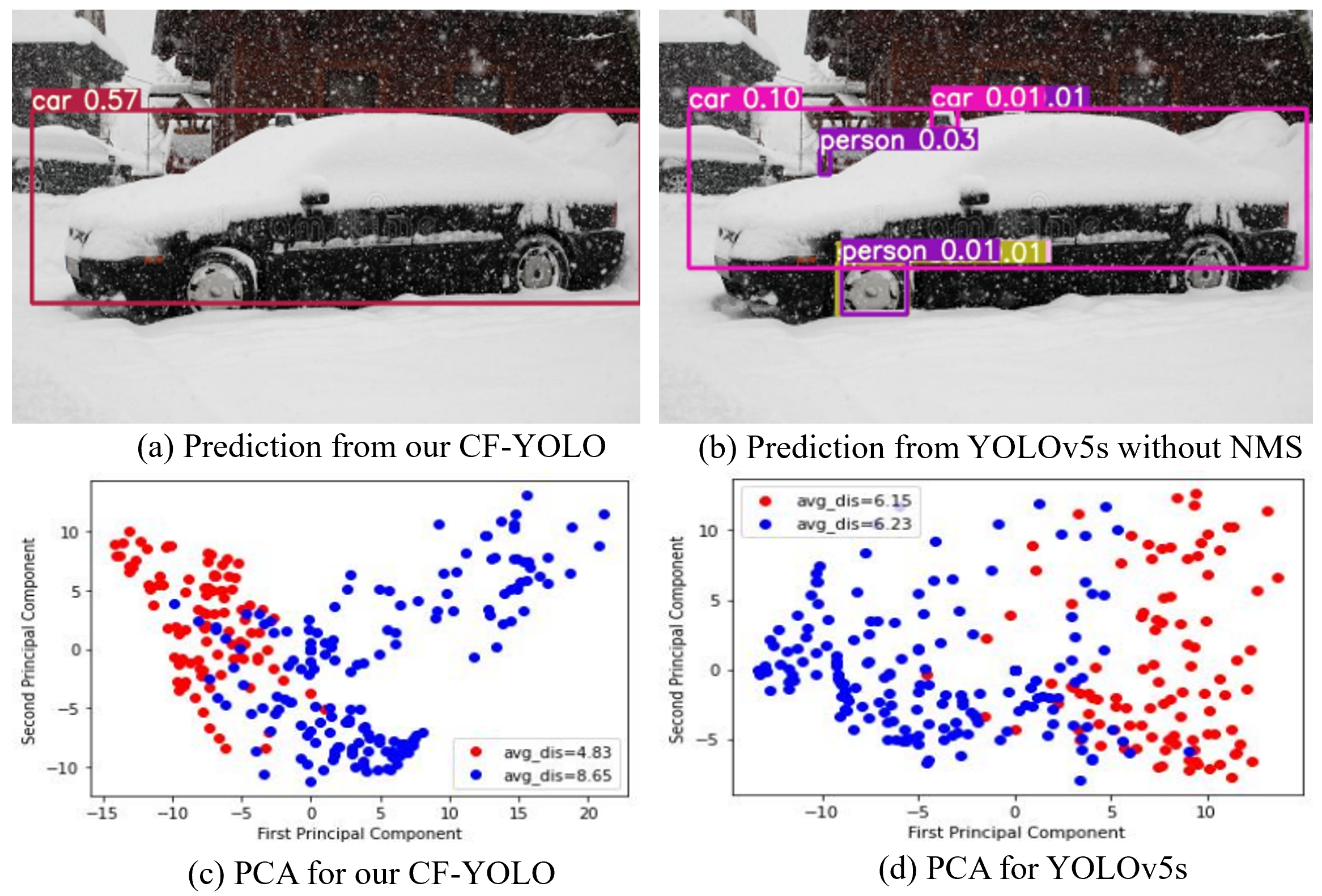}
\caption{PCA study of CF-YOLO and YOLOv5s, the red and blue points indicate the pixels of object and background, respectively.}
\label{pre_ana}
\end{figure}

\subsection{Performance of Detectors on RSOD}
Our CF-YOLO is implemented in PyTorch. All training settings are as same as YOLOv5s (the batch size=32, the SGD optimizer with the momentum of 0.937 and the weight decay of 0.0005, and the learning rate=0.01). We compare our CF-YOLO with different SOTA methods, including YOLOv5s\cite{DBLP:yolov5}, SSD300 \cite{DBLP:conf/eccv/LiuAESRFB16}, EfficientDet D0 and D1 \cite{DBLP:conf/cvpr/TanPL20}.

Furthermore, RSOD is divided into the training, validation and test sets with 1701, 189 and 210 images respectively. To balance the difficulty of each subset, the images are randomly allocated to the subsets.

In order to verify the four-difficulty levels of RSOD, we test on the four levels respectively and the whole dataset. For a fair comparison, all detectors are trained with MSCOCO only. As shown in Table \ref{tab1}, CF-YOLO achieves steady and obvious advantages over SOTAs in each difficulty level and the whole dataset. Besides, as shown in Fig. \ref{prediction}, compared with YOLOv5s, CF-YOLO has a higher confidence in detection results and can reduce missed and false detections. The reason is that the proposed CF block enables the direct interaction of the features at different levels, so that the meaningful information diluted in the high-level features can be recovered. Furthermore, as shown in Table \ref{tab1}, detectors suffer severe degradation on different difficulty levels. This proves that our grading method can accurately grade snow images into different levels.

\begin{table*}
\centering
\scriptsize
\tiny
\caption{Experimental results of different methods before and after training on RSOD. Training on RSOD brings significant improvements to the models' performance. SOTAs are also more robust in snowy scenes by training on RSOD.}
\resizebox{1.8\columnwidth}{!}{
\begin{tabular}{c|cccc}
\Xhline{1.2pt}
Method &AP (val) &AP$_{50}$ (val) &AP (test) &AP$_{50}$ (test) \\ \hline
&\multicolumn{4}{c}{Before training / Training on RSOD (20 epochs)} \\ \hline
YOLOv5s &34.2 / 41.1&49.8 / 62.5&32.7 / 37.9&55.4 / 64.6 \\  \hline
EfficientDet D0 &23.0 / 28.9 &33.3 / 47.4 &26.3 / 33.3 &40.0 / 53.1\\ 
EfficientDet D1 &25.1 / 37.8 &34.5 / 56.8 &27.9 / 40.2 &41.6 / 62.4\\ \hline
SSD300 &23.9 / 34.7&42.4 / 59.8&25.6 / 33.1&46.8 / 59.5 \\ \hline
CF-YOLO (K=1) &\textbf{35.5} / \textbf{41.2} &\textbf{51.8} / \textbf{64.6} &\textbf{35.6} / \textbf{42.4} &\textbf{57.6} / \textbf{70.6}\\ 
CF-YOLO (K=3) &\textbf{38.9} / \textbf{47.5} &\textbf{56.5} / \textbf{71.1} &\textbf{34.0} / \textbf{41.9} &\textbf{58.4} / \textbf{70.6}\\  \hline
\Xhline{1.2pt}
\end{tabular}}
\label{table:RSOD}
\end{table*}

Subsequently, to compare the performance of different methods after training on the RSOD dataset, we train the detectors on RSOD with the MSCOCO pre-trained weights. We train the networks with only 20 epochs, which is enough since RSOD is much smaller than MSCOCO. As shown in Tab. \ref{table:RSOD}, CF-YOLO still outperforms SOTAs a lot on the validation and test sets, which confirms the advantage of CF-YOLO in snow weather.

To further study how CF works, we conduct PCA for the outputs of both our CF block and the aggregation module in YOLOv5s (see Fig. \ref{pre_ana} (c) and (d)): \textbf{the red points indicate the object pixels and the blue points represent the background pixels}. We calculate the average distance to the center of its corresponding cluster. It is easy to find that the object pixels of CF-YOLO are better clustered in the feature space, which demonstrates the efficiency of our method on the snowy scenes.

\subsection{Performance of Detectors on MSCOCO}
To further investigate the generalization ability of CF-YOLO, we train two versions ($K=1$ or $3$) of CF-YOLO from scratch for 300 epochs on MSCOCO. Tab. \ref{table:COCO} shows the comparison between CF-YOLO and SOTAs on MSCOCO. We can see that CF-YOLO with the kernel size of $K=1$ or $K=3$ achieves close results to YOLOv5s. It means our CF-YOLO performs well in snowy weather while still being competitive in normal weather.

\subsection{Computational Complexity}
CF-YOLO (K=3) has much more parameters than YOLOv5s and CF-YOLO (K=1) (see Tab. \ref{table:COCO}), since gOctConv handles feature maps from and to all stages simultaneously, resulting in such the excessively large parameter size. gOctConv takes up most of the parameters in the CF block. The parameter size of one single CF block rises from 2.3M to 8.7M with the increase of the kernel size from 1 to 3. 
Despite the disproportional parameter rising, CF-YOLO (K=3) and CF-YOLO (K=1) consume similar hardware resources to YOLOv5s as we can see their FLOPS are similar. The change of kernel size improves mAP by 0.6\% and 0.4\% on MSCOCO validation set and test set, respectively.
The FPS of CF-YOLO is lower than other methods. This is mainly caused by the feature fusion process, for CF cannot concurrently conduct the different stages. Still, the speed of CF-YOLO is sufficient for possible application scenarios.
\begin{table*}[ht]
\centering
\scriptsize
\caption{Comparison of CF-YOLO with SOTAs on MSCOCO. We conduct the speed test on a single Tesla V100 GPU with a batch size of 1, taking the average speed of 5000 images of COCO val2017.}
\resizebox{1.8\columnwidth}{!}{
\begin{tabular}{c|ccccccc}
\Xhline{1.2pt}
Method &Params &GFLOPS &AP (val) &AP$_{50}$ (val) &AP (test) &AP$_{50}$ (test) &FPS \\ \hline
YOLOv5s &7.3M &17.3 &36.3 &55.3 &36.6 &55.3 &65\\
EfficientDet D0 &3.9M &2.5 &34.3 &- &33.8 &55.2 &63 \\ 
EfficientDet D1 &6.6M &6.1 &40.2 &- &39.6 &58.6 &50 \\ 
\hline
CF-YOLO (K=1) &9.2M &17.4 &35.5 &55.6 &35.8 &55.7 &49\\ 
CF-YOLO (K=3) &22M &17.4 &36.1 &55.8 &36.2 &55.9 &44\\ 
\Xhline{1.2pt}
\end{tabular}}
\label{table:COCO}
\end{table*}

\section{Conclusion}
\label{con}
Adverse weather often creates the visibility problem for the sensors that power automated systems.
While cutting-edge object detectors have
obtained promising results on the datasets captured in normal weather, it is
still non-trivial to detect objects from the low-quality images captured in adverse weather (e.g., snowy weather). They often ignore the latent
information beneficial for detection.
By developing an unsupervised training strategy, we establish a high-quality real-world snow dataset for object detection (RSOD). Considering the degradation of CNN-based detectors on RSOD, we propose cross fusion YOLO (CF-YOLO): a lightweight yet effective objection detector.
The results show that our CF-YOLO not only achieves excellent performance on RSOD, but also is a competitive and lightweight general detector, which will facilitate the outdoor vision systems.

\bibliographystyle{IEEEtran}
\bibliography{ref}

\begin{thebibliography}{10}
\providecommand{\url}[1]{#1}
\csname url@samestyle\endcsname
\providecommand{\newblock}{\relax}
\providecommand{\bibinfo}[2]{#2}
\providecommand{\BIBentrySTDinterwordspacing}{\spaceskip=0pt\relax}
\providecommand{\BIBentryALTinterwordstretchfactor}{4}
\providecommand{\BIBentryALTinterwordspacing}{\spaceskip=\fontdimen2\font plus
\BIBentryALTinterwordstretchfactor\fontdimen3\font minus
  \fontdimen4\font\relax}
\providecommand{\BIBforeignlanguage}[2]{{%
\expandafter\ifx\csname l@#1\endcsname\relax
\typeout{** WARNING: IEEEtran.bst: No hyphenation pattern has been}%
\typeout{** loaded for the language `#1'. Using the pattern for}%
\typeout{** the default language instead.}%
\else
\language=\csname l@#1\endcsname
\fi
#2}}
\providecommand{\BIBdecl}{\relax}
\BIBdecl

\bibitem{Hnewa21}
M.~Hnewa and H.~Radha, ``Multiscale domain adaptive {YOLO} for cross-domain
  object detection,'' \emph{CoRR}, vol. abs/2106.01483, 2021.

\bibitem{HuangLJ21}
S.~Huang, T.~Le, and D.~Jaw, ``Dsnet: Joint semantic learning for object
  detection in inclement weather conditions,'' \emph{{IEEE} Trans. Pattern
  Anal. Mach. Intell.}, vol.~43, no.~8, pp. 2623--2633, 2021.

\bibitem{tgrs/GongXTSXDL20}
Y.~Gong, Z.~Xiao, X.~Tan, H.~Sui, C.~Xu, H.~Duan, and D.~Li, ``Context-aware
  convolutional neural network for object detection in {VHR} remote sensing
  imagery,'' \emph{{IEEE} Trans. Geosci. Remote. Sens.}, vol.~58, no.~1, pp.
  34--44, 2020.

\bibitem{DBLP:journals/corr/RenHG015}
S.~Ren, K.~He, R.~B. Girshick, and J.~Sun, ``Faster {R-CNN:} towards real-time
  object detection with region proposal networks,'' \emph{CoRR}, vol.
  abs/1506.01497, 2015.

\bibitem{tgrs/ZhangYFL19}
Y.~Zhang, Y.~Yuan, Y.~Feng, and X.~Lu, ``Hierarchical and robust convolutional
  neural network for very high-resolution remote sensing object detection,''
  \emph{{IEEE} Trans. Geosci. Remote. Sens.}, vol.~57, no.~8, pp. 5535--5548,
  2019.

\bibitem{DBLP:journals/tgrs/GuoYYTHL21a}
J.~Guo, J.~Yang, H.~Yue, H.~Tan, C.~Hou, and K.~Li, ``Rsdehazenet: Dehazing
  network with channel refinement for multispectral remote sensing images,''
  \emph{{IEEE} Trans. Geosci. Remote. Sens.}, vol.~59, no.~3, pp. 2535--2549,
  2021.

\bibitem{DengDSLL009}
J.~Deng, W.~Dong, R.~Socher, L.~Li, K.~Li, and L.~Fei{-}Fei, ``Imagenet: {A}
  large-scale hierarchical image database,'' in \emph{{CVPR} 2009}, 2009, pp.
  248--255.

\bibitem{DBLP:conf/eccv/LinMBHPRDZ14}
T.~Lin, M.~Maire, S.~J. Belongie, J.~Hays, P.~Perona, D.~Ramanan,
  P.~Doll{\'{a}}r, and C.~L. Zitnick, ``Microsoft {COCO:} common objects in
  context,'' in \emph{{ECCV} 2014}, 2014, pp. 740--755.

\bibitem{DBLP:journals/tgrs/CaoFXM22}
X.~Cao, X.~Fu, C.~Xu, and D.~Meng, ``Deep spatial-spectral global reasoning
  network for hyperspectral image denoising,'' \emph{{IEEE} Trans. Geosci.
  Remote. Sens.}, vol.~60, pp. 1--14, 2022.

\bibitem{DBLP:journals/tgrs/ZhangLZ19}
G.~Zhang, S.~Lu, and W.~Zhang, ``Cad-net: {A} context-aware detection network
  for objects in remote sensing imagery,'' \emph{{IEEE} Trans. Geosci. Remote.
  Sens.}, vol.~57, no.~12, pp. 10\,015--10\,024, 2019.

\bibitem{DBLP:journals/corr/abs-2103-11298}
K.~Zhang, R.~Li, Y.~Yu, W.~Luo, C.~Li, and H.~Li, ``Deep dense multi-scale
  network for snow removal using semantic and geometric priors,'' \emph{CoRR},
  2021.

\bibitem{DBLP:conf/cvpr/LinDGHHB17}
T.~Lin, P.~Doll{\'{a}}r, R.~B. Girshick, K.~He, B.~Hariharan, and S.~J.
  Belongie, ``Feature pyramid networks for object detection,'' in \emph{{CVPR}
  2017}, 2017, pp. 936--944.

\bibitem{DBLP:conf/cvpr/GirshickDDM14}
R.~B. Girshick, J.~Donahue, T.~Darrell, and J.~Malik, ``Rich feature
  hierarchies for accurate object detection and semantic segmentation,'' in
  \emph{{CVPR} 2014}, 2014, pp. 580--587.

\bibitem{DBLP:conf/iccv/Girshick15}
R.~B. Girshick, ``Fast {R-CNN},'' in \emph{2015 {IEEE} International Conference
  on Computer Vision, {ICCV} 2015, Santiago, Chile, December 7-13, 2015}.\hskip
  1em plus 0.5em minus 0.4em\relax {IEEE} Computer Society, 2015, pp.
  1440--1448.

\bibitem{DBLP:conf/nips/DaiLHS16}
J.~Dai, Y.~Li, K.~He, and J.~Sun, ``{R-FCN:} object detection via region-based
  fully convolutional networks,'' in \emph{{NIPS} 2016}, 2016, pp. 379--387.

\bibitem{DBLP:conf/cvpr/PangCSFOL19}
J.~Pang, K.~Chen, J.~Shi, H.~Feng, W.~Ouyang, and D.~Lin, ``Libra {R-CNN:}
  towards balanced learning for object detection,'' in \emph{{CVPR} 2019},
  2019, pp. 821--830.

\bibitem{DBLP:conf/cvpr/RedmonDGF16}
J.~Redmon, S.~K. Divvala, R.~B. Girshick, and A.~Farhadi, ``You only look once:
  Unified, real-time object detection,'' in \emph{{CVPR} 2016}, 2016, pp.
  779--788.

\bibitem{DBLP:journals/corr/abs-2004-10934}
A.~Bochkovskiy, C.~Wang, and H.~M. Liao, ``Yolov4: Optimal speed and accuracy
  of object detection,'' \emph{CoRR}, 2020.

\bibitem{DBLP:yolov5}
glenn jocher~et al., ``yolov5. https://github.com/ultralytics/yolov5,'' 2021.

\bibitem{DBLP:conf/eccv/LiuAESRFB16}
W.~Liu, D.~Anguelov, D.~Erhan, C.~Szegedy, S.~E. Reed, C.~Fu, and A.~C. Berg,
  ``{SSD:} single shot multibox detector,'' in \emph{{ECCV} 2016}, 2016, pp.
  21--37.

\bibitem{DBLP:conf/iccv/LinGGHD17}
T.~Lin, P.~Goyal, R.~B. Girshick, K.~He, and P.~Doll{\'{a}}r, ``Focal loss for
  dense object detection,'' in \emph{{ICCV} 2017}, 2017, pp. 2999--3007.

\bibitem{DBLP:conf/cvpr/TanPL20}
M.~Tan, R.~Pang, and Q.~V. Le, ``Efficientdet: Scalable and efficient object
  detection,'' in \emph{{CVPR} 2020}, 2020, pp. 10\,778--10\,787.

\bibitem{DBLP:conf/eccv/CarionMSUKZ20}
N.~Carion, F.~Massa, G.~Synnaeve, N.~Usunier, A.~Kirillov, and S.~Zagoruyko,
  ``End-to-end object detection with transformers,'' in \emph{{ECCV} 2020},
  2020, pp. 213--229.

\bibitem{eccv/SindagiOYP20}
V.~A. Sindagi, P.~Oza, R.~Yasarla, and V.~M. Patel, ``Prior-based domain
  adaptive object detection for hazy and rainy conditions,'' in \emph{ECCV},
  ser. Lecture Notes in Computer Science, A.~Vedaldi, H.~Bischof, T.~Brox, and
  J.~Frahm, Eds., vol. 12359, 2020, pp. 763--780.

\bibitem{DBLP:journals/tcsv/JawHK21}
D.~Jaw, S.~Huang, and S.~Kuo, ``Desnowgan: An efficient single image snow
  removal framework using cross-resolution lateral connection and gans,''
  \emph{{IEEE} Trans. Circuits Syst. Video Technol.}, pp. 1342--1350, 2021.

\bibitem{DBLP:journals/corr/abs-2112-06451}
D.~Liang, L.~Li, M.~Wei, S.~Yang, L.~Zhang, W.~Yang, Y.~Du, and H.~Zhou,
  ``Semantically contrastive learning for low-light image enhancement,''
  \emph{CoRR}, vol. abs/2112.06451, 2021.

\bibitem{DBLP:journals/tip/LiuJHH18}
Y.~Liu, D.~Jaw, S.~Huang, and J.~Hwang, ``Desnownet: Context-aware deep network
  for snow removal,'' \emph{{TIP} 2018}, pp. 3064--3073, 2018.

\bibitem{DBLP:conf/cvpr/LiuQQSJ18}
S.~Liu, L.~Qi, H.~Qin, J.~Shi, and J.~Jia, ``Path aggregation network for
  instance segmentation,'' in \emph{{CVPR} 2018}, 2018, pp. 8759--8768.

\bibitem{DBLP:conf/cvpr/GhiasiLL19}
G.~Ghiasi, T.~Lin, and Q.~V. Le, ``{NAS-FPN:} learning scalable feature pyramid
  architecture for object detection,'' in \emph{{CVPR} 2019}, 2019, pp.
  7036--7045.

\bibitem{DBLP:journals/corr/abs-1911-09516}
S.~Liu, D.~Huang, and Y.~Wang, ``Learning spatial fusion for single-shot object
  detection,'' \emph{CoRR}, vol. abs/1911.09516, 2019.

\bibitem{DBLP:conf/iccv/Chen0XYKRYF19}
Y.~Chen, H.~Fan, B.~Xu, Z.~Yan, Y.~Kalantidis, M.~Rohrbach, S.~Yan, and
  J.~Feng, ``Drop an octave: Reducing spatial redundancy in convolutional
  neural networks with octave convolution,'' in \emph{{ICCV} 2019}, 2019, pp.
  3434--3443.

\bibitem{DBLP:conf/eccv/GaoTCLCY20}
S.~Gao, Y.~Tan, M.~Cheng, C.~Lu, Y.~Chen, and S.~Yan, ``Highly efficient
  salient object detection with 100k parameters,'' in \emph{{ECCV} 2020}, 2020,
  pp. 702--721.

\bibitem{DBLP:conf/cvpr/WangLWCHY20}
C.~Wang, H.~M. Liao, Y.~Wu, P.~Chen, J.~Hsieh, and I.~Yeh, ``Cspnet: {A} new
  backbone that can enhance learning capability of {CNN},'' in \emph{{CVPR}
  Workshops 2020}, 2020, pp. 1571--1580.

\bibitem{2011Deep}
X.~Glorot, A.~Bordes, and Y.~Bengio, ``Deep sparse rectifier neural networks,''
  \emph{Journal of Machine Learning Research}, vol.~15, pp. 315--323, 2011.

\bibitem{maas2013rectifier}
A.~L. Maas, A.~Y. Hannun, A.~Y. Ng \emph{et~al.}, ``Rectifier nonlinearities
  improve neural network acoustic models,'' in \emph{Proc. icml}, vol.~30,
  no.~1.\hskip 1em plus 0.5em minus 0.4em\relax Citeseer, 2013, p.~3.

\end{thebibliography}

\vfill

\end{document}